\documentclass[letterpaper, 10 pt, conference]{ieeeconf}  % Comment this line out
\IEEEoverridecommandlockouts         % This command is only
\usepackage{balance} %even columns on 
\usepackage{xcolor,calc}
\usepackage[pdftex]{graphicx}
\usepackage{amsmath} % assumes amsmath package installed
\usepackage{amssymb}  % assumes amsmath package installed
\usepackage[noadjust]{cite}
\usepackage{color}
\usepackage{enumerate}
\usepackage{multirow}
\usepackage{mathtools}
\usepackage{booktabs}
\usepackage{epstopdf}
\usepackage{algcompatible}
\usepackage[linesnumbered,ruled,vlined]{algorithm2e}

\usepackage[normalem]{ulem}%%%%%%%%%
\usepackage{marvosym}
\usepackage{url}
\usepackage{bm}

\usepackage{booktabs}

\newtheorem{remark}{Remark}

\SetKwInput{KwInit}{Initialization}                % Set the Input
\SetKwInput{KwInput}{Input}
\SetKwInput{KwOutput}{Output}
\makeatletter
\newsavebox\myboxA
\newsavebox\myboxB
\newlength\mylenA

\newcommand*\xoverline[2][0.75]{%
    \sbox{\myboxA}{$\m@th#2$}%
    \setbox\myboxB\null% Phantom box
    \ht\myboxB=\ht\myboxA%
    \dp\myboxB=\dp\myboxA%
    \wd\myboxB=#1\wd\myboxA% Scale phantom
    \sbox\myboxB{$\m@th\overline{\copy\myboxB}$}%  Overlined phantom
    \setlength\mylenA{\the\wd\myboxA}%   calc width diff
    \addtolength\mylenA{-\the\wd\myboxB}%
    \ifdim\wd\myboxB<\wd\myboxA%
       \rlap{\hskip 0.5\mylenA\usebox\myboxB}{\usebox\myboxA}%
    \else
        \hskip -0.5\mylenA\rlap{\usebox\myboxA}{\hskip 0.5\mylenA\usebox\myboxB}%
    \fi}
\makeatother

\usepackage{tabularx}
\usepackage{booktabs}

\newtheorem{definition}{Definition}
\newtheorem{assumption}{Assumption}

\makeatletter
\let\NAT@parse\undefined
\makeatother
\usepackage{hyperref}

\title{\LARGE \bf Learning to Play Cup-and-Ball with Noisy Camera Observations}

\author{Monimoy Bujarbaruah$^{\star, 1}$, Tony Zheng$^{\star, 1}$, Akhil Shetty$^{\star, 1}$, Martin Sehr, Francesco Borrelli$^{1}$ % <-this % stops a space
\thanks{$^1$The authors are with UC Berkeley, USA, and author Martin Sehr is with Siemens Corporate Technology, USA; E-mails: \tt\scriptsize{\{monimoyb, tony\_zheng, shetty.akhil, fborrelli\}@berkeley.edu,} \tt\scriptsize{martin.sehr@siemens.com.}}
\thanks{$^\star$ These authors contributed equally to this work.
}
}

\begin{document}

\maketitle
  \thispagestyle{empty}
\pagestyle{empty}

%%%%%%%%%%%%%%%%%%%%%%%%%%%%%%%%%%%%
\begin{abstract}
Playing the cup-and-ball game is an intriguing task for robotics research since it abstracts important problem characteristics including system nonlinearity, contact forces and precise positioning as terminal goal. In this paper, we present a learning model based control strategy for the cup-and-ball game, where a Universal Robots UR5e manipulator arm learns to catch a ball in one of the cups on a Kendama.  
Our control problem is divided into two sub-tasks, namely $(i)$ swinging the ball up in a constrained motion, and $(ii)$ catching the free-falling ball. The swing-up trajectory is computed \emph{offline}, and applied in open-loop to the arm. Subsequently, a convex optimization problem is solved \emph{online} during the ball's free-fall to control the manipulator and catch the ball. The controller utilizes noisy position feedback of the ball from an Intel RealSense D435 depth camera. We propose a novel iterative framework, where data is used to learn the support of the camera noise distribution iteratively in order to update the control policy.  The probability of a catch with a fixed policy is computed empirically with a user specified number of roll-outs. Our design guarantees that probability of the catch increases in the limit, as the learned support nears the true support of the camera noise distribution. High-fidelity Mujoco simulations and preliminary experimental results support our theoretical analysis (\href{https://www.dropbox.com/s/ih4a5uld197rd42/Kendama2020_Successful_Catch.mp4?dl=0}{\textbf{video link}} -- \href{https://github.com/MPC-Berkeley/kendama}{\textbf{GitHub link}}).
\end{abstract}
%%%%%%%%%%%%%%%%%%%%%%%%%%%%%%%%%%%%%%%%%%%%%%%

\section{Introduction}
Kendama is the Japanese version of the classic cup-and-ball game, which consists of a handle, a pair of cups, and a ball, which are all connected by a string. Playing the cup-and-ball game is a task commonly considered in robotics research \cite{Nemec_Ude, Kober_Peters, Miyamoto_7, Sakaguchi_Miyazaki, Namiki_Itoi, Schwab_7, Senoo_3, li2020robot}, where approaches ranging from classical PD control to reinforcement learning have been utilized to solve the task. The model-based approaches among the above typically decompose the task into two sub-tasks, namely $(i)$ performing a swing-up of the ball when the string is taut, and $(ii)$ catching the ball during its free-fall. The models of the joint system considered for both sub-tasks are different, thus resulting in hybrid control design for the robotic manipulator. The key drawbacks in such existing approaches are namely the need for expert demonstrations, and the lack of guarantees of operating constraint satisfaction and obtaining catches under modeling uncertainty and sensing errors.

In this paper, we propose a fully physics driven model-based hybrid approach for control design. The controller guarantees a \emph{constrained} motion, while accounting for our best \emph{estimates} of  uncertainty in the system model and sensing errors. We use a mixed open-loop and closed-loop control design, motivated by works such as \cite{Hansen1996ReinforcementOLCL, schaalOLCL, thrunOLCL}. First, the swing-up phase is designed offline and then an \emph{open-loop} policy is applied to the robotic manipulator. We use a \emph{cart with inverted pendulum model} of the cup-and-ball joint system for swing-up policy design. For this phase, as we solve a constrained finite horizon non-convex optimization problem, we only consider a \emph{nominal} disturbance-free model of the system. The swing-up trajectory is thus designed to ensure that the predicted difference in positions of the ball and the cup vanishes at a future time once the nominal terminal swing-up state is reached and the cup is held fixed. 

After a swing-up, we switch to online closed-loop control synthesis once the ball starts its free-fall. We consider presence of only a camera that takes noisy measurements of the ball's position at every time step. We design the feedback controller in the manipulator's end-effector \cite{MLS} space. This results in a Linear Time Invariant (LTI) model for the evolution of the difference between the cup and the ball's positions, thus allowing us to solve convex optimization problems online for control synthesis. In order to guarantee a catch by minimizing the position difference, it is also crucial to ensure that during the free-fall of the ball, the control actions to the manipulator do not yield a configuration where the string is taut, despite uncertainty in the model and noise in camera position measurements. Uncertainty in the LTI model primarily arises from low level controller mismatches in the manipulator hardware, and an upper bound of this uncertainty is assumed \emph{known}. Bounds on the measurement noise induced by the camera are assumed unknown. This paper presents a method to increase the  \emph{probability} of a catch,  as the estimate of the support of camera measurement noise distribution is updated. Our contributions are summarized as: 
\begin{itemize}
    \item \emph{Offline}, before the feedback control of the manipulator, we design a swing-up trajectory for the nominal cup-and-ball system that plans the motion of the ball to a state from which a catch control is initiated.  

    \item Using the notion of \emph{Confidence Support} from \cite{bujarbaruah2019learning} which is guaranteed to contain the true support of the camera measurement noise with a specified probability, we use \emph{online} robust feedback control  for enforcing bounds on the probability of failed catches. 
    
    \item With high-fidelity Mujoco simulations and preliminary physical experiments we demonstrate that the manipulator gets better at catching the ball as the support of the camera measurement noise is learned  and as the Confidence Support and closed-loop policy are updated.
\end{itemize}
\section{Generating A Swing-up Trajectory}\label{sec:swing_up}
\begin{figure}[h]
    \centering
    \includegraphics[width=0.7\columnwidth]{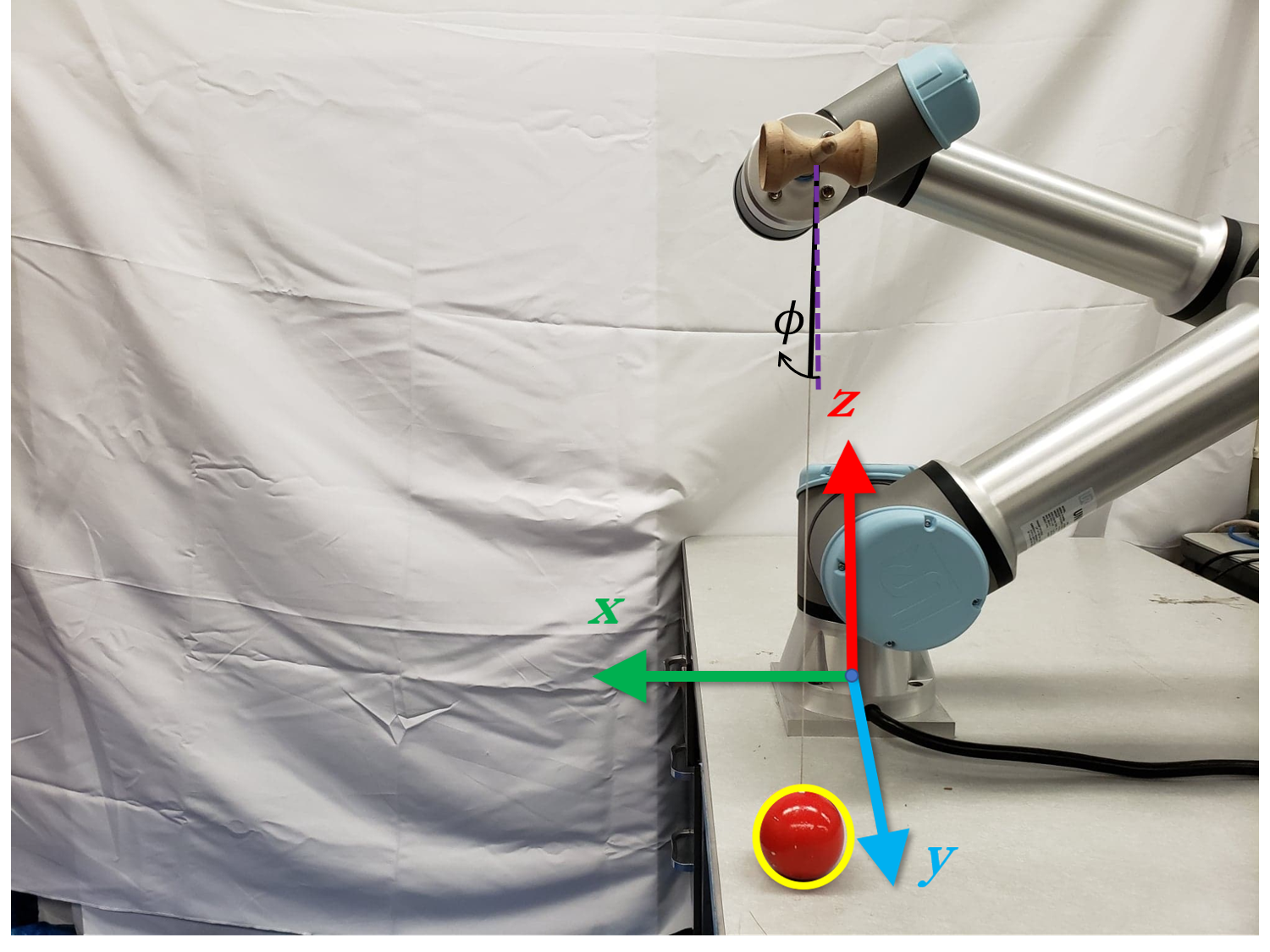}
    \caption{Manipulator with Kendama along with coordinate frame.}
    \vspace{-8pt}
    \label{fig:kendama_intro}
\end{figure}

The swing-up phase begins with the arm in the home position such that the ball is hanging down at an angle of $0$ radians from the vertical plumb line, as seen in Fig.~\ref{fig:kendama_intro}. 

\subsection{System Modeling}
We model the system such that the cup is a planar cart with point-mass $m_c$ and the ball acts as a rigid pendulum (mass $m_b$ and radius $r$) attached to the cup. Assuming planar $xz$-motion of the ball, we derive the Lagrange equations of motion \cite{MLS} with three generalized coordinates $\mathbf{q}(t)=(x^\mathrm{cup}(t),z^\mathrm{cup}(t),\phi(t))$, which denote the $x$ position of the cup, $z$ position of the cup, and swing angle of the ball with respect to the plumb line of the cup respectively at any time $t \geq 0$. We reduce the equations to the general \emph{nominal} form
\begin{equation}\label{lagrange}
    M(\mathbf{q}(t))\ddot{\mathbf{q}}(t) + C(\mathbf{q}(t),\dot{\mathbf{q}(t)})\dot{\mathbf{q}(t)} + G(\mathbf{q}(t)) = F(t),~\forall t \geq 0,
\end{equation}
where $M(\mathbf{q}(t))$ is the inertia matrix, $C(\mathbf{q}(t),\dot{\mathbf{q}}(t))$ is the Coriolis matrix, $G(\mathbf{q}(t))$ is the gravity matrix, and $F(t)$ is the external input force at time $t$. Here $\dot{\mathbf{q}}(t)$ denotes the velocity of the cup and the angular velocity of the ball, and $\ddot{\mathbf{q}}(t)$ denotes the acceleration of the cup and the angular acceleration of the ball at any time $t \geq 0$. System \eqref{lagrange} in state-space form is
\begin{equation}\label{statespacelagrange}
    \dot{\bar{x}}(t) = f(\bar{x}(t), F(t)),
\end{equation}
where nominal state $\bar{x}(t) = [\mathbf{q}^\top(t), \dot{\mathbf{q}}^\top(t)]^\top \in \mathbb{R}^6$ for all time $t \geq 0$.
% , and the system dynamics matrix $f(\cdot, \cdot)$ is given in the Appendix.  
%%%%%%%%%%%%%%%%%%%%%%%%%%%%%%%%%%%%%
\subsection{Optimization Problem}\label{ssec:swingOpt}
We discretize system \eqref{statespacelagrange} with one step  Euler discretization and a sampling time of $T_s = 100$Hz. The discrete time system can then be written as 
\begin{align*}
    \bar{x}_{i+1} = \bar{x}_i + T_s f(\bar{x}_i, F_i) = f_d(\bar{x}_i, F_i),~\forall i \in \{0,1,\dots\},
\end{align*}
where $a_i$ denotes the sampled time version of continuous variable $a(t)$. To generate a force input sequence for the swing-up, we solve a constrained optimal control problem over a finite planning horizon of length $N$, given by:
\begin{equation}\label{eq:swinger_eq}	\begin{array}{clll}
		\displaystyle\min_{F_0,\dots, F_{N-1}} & \displaystyle\sum\limits_{i=0}^{N-1} \bar{x}_i^\top Q_s \bar{x}_i + F_i^\top R_s F_i  \\[1ex]
		\text{s.t.,}  & \bar{x}_{i+1} = f_d(\bar{x}_i,F_i),\\
		& \bar{x}_i \in \mathcal{X},  F_i \in \mathcal{F},\\
        & \bar{x}_0 = x_\mathrm{init},\\
        & \bar{x}_{N} = x_\mathrm{f},~i = 0,1,\dots,(N-1),
	\end{array}
\end{equation}
where weight matrices $Q_s, R_s \succ 0$, and constraint set $\mathcal{X}$ is chosen such that the ball remains within the reach of the UR5e manipulator. Initial state $x_\mathrm{init}$ is known in the configuration as shown in Fig.~\ref{fig:kendama_intro}. Due to the nonlinear dynamics $f_d(\cdot, \cdot)$, the optimization problem \eqref{eq:swinger_eq} is non-convex. Moreover, typically a long horizon length $N$ is required. Hence, we solve \eqref{eq:swinger_eq} \emph{offline} and apply the computed input sequence $\mathbf{F}^\star = [F^\star_0, F^\star_1, \dots, F^\star_{N-1}]$ in open-loop to the manipulator.

\subsection{Terminal Conditions of the Swing-Up}\label{bxfsec}
\subsubsection*{Predicted Behaviour}
The nominal terminal state $x_\mathrm{f}$ in \eqref{eq:swinger_eq} is selected such that the ball is swinging to $\phi = 2.44$ rad with an angular velocity of $\dot{\phi} = 4.18$ rad/s. At these values, the string is calculated to lose tension and the ball begins free-fall. The chosen value of $x_\mathrm{f}$ ensures that the predicted difference in positions of the ball and the cup (both modeled as point masses) vanishes at a future time, if the cup were held fixed and the ball's motion is predicted under free-fall. 

\subsubsection*{Actual Behaviour}
When considering the nominal system \eqref{lagrange}, we have ignored the presence of uncertainties. Such uncertainties may arise due to our simplifying assumptions such as: $(i)$ the string is mass-less so the swing angle is only affected by the ball and cup masses, $(ii)$ there are no frictional and aerodynamic drag forces to hinder the conservation of kinetic and potential energy of the system, $(iii)$ the cup mass is decoupled from the mass of the manipulator, and $(iv)$ there is no mismatch of control commands from the low level controller of the manipulator and $F$. Due to such uncertainties, realized states $x_i$ for $i \in \{0,1,\dots,N\}$ do not \emph{exactly} match their nominal counterparts. 

\begin{figure}[h]
    \centering
    \includegraphics[width=0.8\columnwidth]{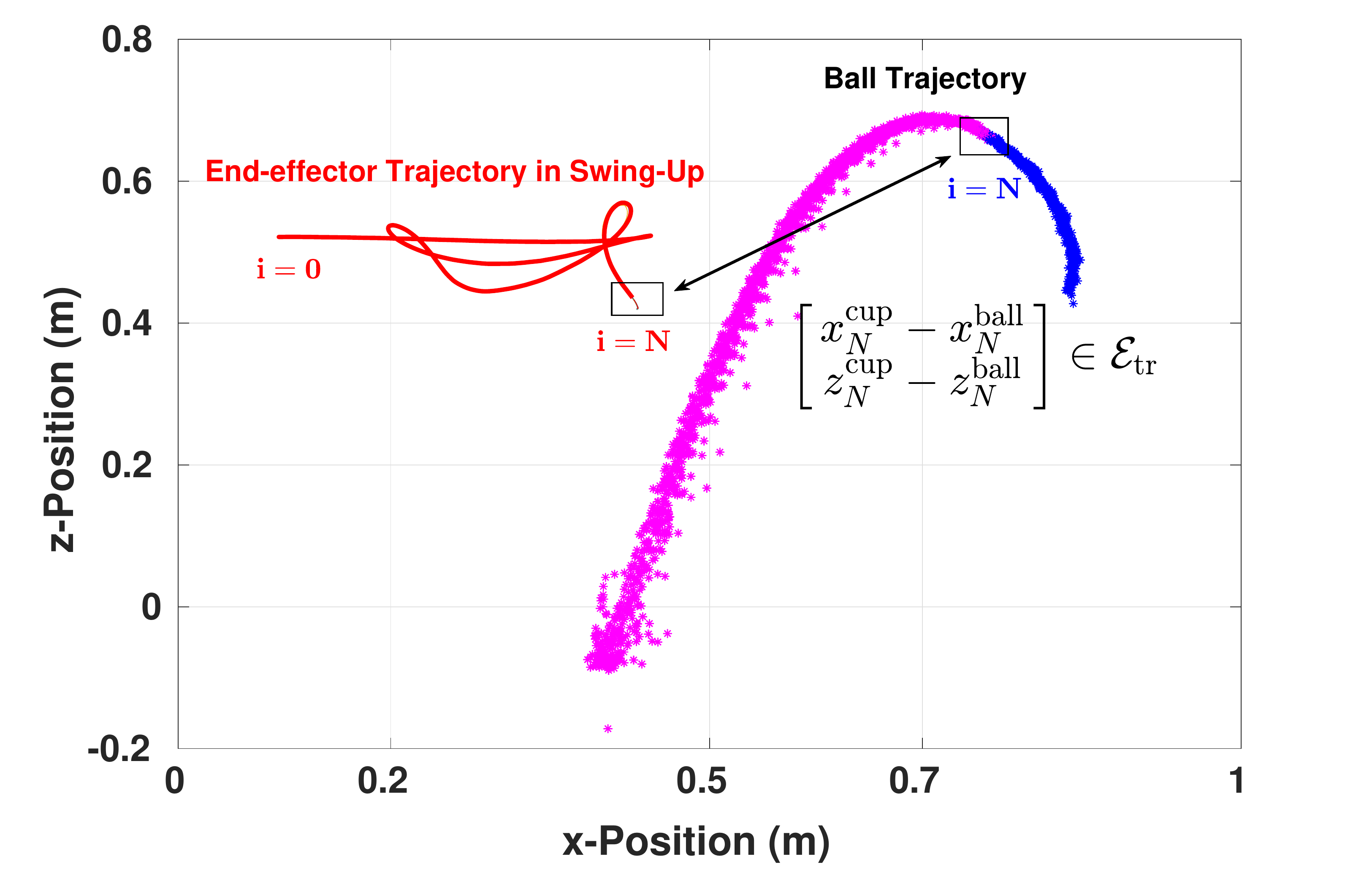}
    \caption{Start of catch phase (i.e., $i=N$) for 100 trajectories. Red line indicates the trajectory of the cup/end-effector during swing-up. Blue dots indicate ball positions during swing-up and pink dots indicate a position after catch phase is started. Closed-loop control begins when the relative position is in $\mathcal{E}_\mathrm{tr}$.}
    \vspace{-8pt}
    \label{fig:roll_out}
\end{figure}

A set of 100 measured roll-out trajectories of the ball after the swing-up are shown in Fig.~\ref{fig:roll_out} for a fixed open-loop input sequence $\mathbf{F}^\star$.
We see from Fig.~\ref{fig:roll_out} that after $N$ time steps of swing-up, the ball and the cup arrive at positions where their relative position is in a set $\mathcal{E}_\mathrm{tr}$. 
A key assumption of well posedness will be imposed on this set in Section~\ref{wellsec} in order for our subsequent feedback control policy to deliver a catch in experiments. 
%%%%%%%%%%%%%%%%%%%%%%%%%%%%%%%%%%%%%%%%%%%%%%
\section{Designing Feedback Policy In Catch Phase}\label{sec:catch_learn}
For the catch phase we start the time index $t=0$ where the swing up ends, i.e., $i=N$. There are two main challenges during the design of the feedback controller, namely $(i)$ position measurements of the ball from a noisy camera, and $(ii)$ presence of mismatch between desired control actions and corresponding low level controller commands. 
% We make the following assumption.
\begin{assumption}\label{assump:end_eff_acc}
We assume that the UR5e end-effector gives an accurate estimate of its own position. The assumption is based on precision ranges provided in \cite{eseries}.
\end{assumption}

\subsection{Problem Formulation}\label{ssec:feedback_for}
During free-fall of the ball we design our feedback controller for the manipulator position \emph{only} in end-effector space, with desired velocity of the end-effector as our control input. The joint ball and end-effector system in one trial can be modeled as a single integrator as:
\begin{subequations}\label{eq:unc_system}
\begin{align}
e_{t+1} & = A e_t + B  u_t + w_t(e_t,  u_t), \\
 y_t & = e_t + v_t, \label{e0here}
\end{align}
\end{subequations}
with error states and inputs (i.e., relative position and velocity)
\begin{align*}
& e_t = \begin{bmatrix} x^{\mathrm{cup}}_t - x^\mathrm{ball}_t \\ z^\mathrm{cup}_t - z^\mathrm{ball}_t \end{bmatrix},~  u_t = \begin{bmatrix} v^{\mathrm{cup}}_{x,t} - v^\mathrm{ball}_{x,t} \\ v^\mathrm{cup}_{z,t} - v^\mathrm{ball}_{z,t} \end{bmatrix}, 
\end{align*}
where $w_t(e_t,  u_t) \in \mathbb{W}_m \subset \mathbb{R}^2$, is a bounded uncertainty which arises due to the discrepancy between $(i)$ the predicted and the actual velocity of the ball at any given time step\footnote{we use the camera position information for ball's velocity estimation}, and $(ii)$ the commanded and the realized velocities of the end-effector, primarily due to the low level controller delays and limitations. System dynamics matrices $A = {I}_2$ and $B = dt\cdot {I}_2$ are known, where ${I}_d$ denotes the identity matrix of size $d$, and sampling time $dt = 0.01$ second. We assume an outer approximation $\mathbb{W}$ to the set $\mathbb{W}_m$, i.e., $\mathbb{W}_m \subseteq \mathbb{W}$ is known, and is a polytope. We consider noisy measurements of states due to the noise in camera position measurements, corrupted by $v_t \stackrel{\mathrm{i.i.d.}}{\sim} \mathcal{P}$, with $\mathrm{Supp}(\mathcal{P}) = \mathbb{V}$, where $\mathrm{Supp}(\cdot)$ denotes the support of a distribution. We assume $\mathbb{V}$ is \emph{not} exactly known. 

% Since only noisy measurements $y_0 \in \mathcal{E}_\mathrm{tr}$ of state $e_0$ are available (see Fig.~\ref{fig:roll_out}), 
% in \eqref{e0here}, we consider $e_0 \in \mathcal{E}_\mathrm{init}$, where we have: 
% \begin{align}\label{bxyxrelation}
%       \mathcal{E}_\mathrm{init} = y_0  \oplus (-\mathbb{V}).
% \end{align}
Using the set $\mathcal{E}_\mathrm{tr}$ (see Fig.~\ref{fig:roll_out}), a set $\mathcal{E}$ containing the origin where the string is not taut and \eqref{eq:unc_system} is valid can then be chosen. We choose:
\begin{align}\label{act_esxf}
    & \gamma^{(i)} = \Vert \mathrm{vert}^{(i)}( \mathcal{E}_\mathrm{tr}) \Vert_\infty ,~i \in \{1,2\}, \nonumber \\
    & \mathcal{E} = \{x: -\gamma \leq x \leq \gamma\},~\gamma = [\gamma^{(1)}, \gamma^{(2)}]^\top,
\end{align}
where $\mathrm{vert}^{(i)}(\mathcal{A})$ denotes $i^\mathrm{th}$ row of all the vertices of the polytope $\mathcal{A}$, and $\Vert \cdot \Vert$ denotes the vector norm. This ensures 
\begin{align}\label{e0et}
    e_0 \in \mathcal{E}_\mathrm{tr} \implies e_0 \in \mathcal{E}.  
\end{align}

% Consequently, the difference between the cup and the ball's corresponding positions is contained in a set denoted by $\mathcal{E}_S(x_\mathrm{f})$. Certain key assumptions will be imposed on  this set $\mathcal{E}_S(x_\mathrm{f})$, in order for our subsequent feedback control policy to deliver a catch in experiments. These are introduced in Section~\ref{sec:iter_cftoc}.  

As \eqref{e0et} holds true, we impose state and input constraints for all time steps $t \geq 0$ as given by:
\begin{align}\label{eq:constraints_nominal}
	e_t \in \mathcal{E},~ u_t 
	\in \mathcal{U},
\end{align}
where set $\mathcal{U}$ is a polytope.
% \section{Robust Estimation Bounds}\label{sec:rob_est}
% At time $t=0$ of any $j^\mathrm{th}$ iteration, we assume a prior on the bounds of $x^{1;J}_0$, that is $x^{1:J}_0 \in \hat{\mathbb{X}}_0$. Our aim is to obtain robust bounds on states along the horizon, as we gather measurements. These bounds are then used to obtain robust bounds on noise $\hat{\mathbb{V}}^j_t$ along the iteration, with the guarantee that $\hat{\mathbb{V}}^j_t \subset \mathbb{V}$ for all $t \in \{0,1,\dots,T\}$ and for all $j \in [J]$. The update equations are:
% \begin{subequations}\label{eq:estim_update}
% \begin{align}
%     \hat{\mathbb{X}}^{j,\mathrm{ pos}}_t & = \hat{\mathbb{X}}^{j,\mathrm{ pri}}_t \cap (y^j_t - \mathbb{V}_\mathrm{out}),\\
%     \hat{\mathbb{V}}^j_t & = \hat{\mathbb{X}}^{j,\mathrm{pos}}_t \ominus y^j_t,\\
%     \hat{\mathbb{X}}^{j,\mathrm{ pri}}_{t+1} & = A \hat{\mathbb{X}}^{j,\mathrm{ pos}}_t + Bu_t. 
% \end{align}
% \end{subequations}
% After the end of the $j^\mathrm{th}$ iteration we form a set
% \begin{align}\label{eq:control_noise_bounds}
%   \hat{\mathbb{V}}^{j+1, \mathrm{traj}} = \mathrm{conv}(\hat{\mathbb{V}}^{j,\mathrm{traj}} \cup (\cup_{t=0}^{T}\hat{\mathbb{V}}^j_t)) \subset \mathbb{V}_\mathrm{out}.
% \end{align}
% These sets are used for controller design.
We formulate the following finite horizon robust optimal control problem for feedback control design:
\begin{equation}\label{eq:generalized_InfOCP}
	\begin{array}{clll}
% 		\hspace{0cm} V^{\star}(e_0) = \\ [1ex]
		\displaystyle\min_{ u_0,  u_1(\cdot),\ldots} & \displaystyle\sum\limits_{t=0}^{T-1} \ell \left( \bar{e}_t,  u_t\left(\bar{e}_t\right) \right) + Q(\bar{e}_T) \\[1ex]
		\text{s.t.,}  & e_{t+1} = A e_t + B  u_t(e_t) + w_t(e_t,  u_t),\\
		& \bar{e}_{t+1} = A \bar{e}_t + B u_t(\bar{e}_t),\\
		& y_t = e_t + v_t,
		\\[1ex]
		& e_t \in \mathcal{E},  u_t(e_t) \in \mathcal{U},\\[1ex]
		&\forall w_t(e_t,  u_t) \in \mathbb{W},~ \forall v_t \in \mathbb V,\ \\[1ex]
		&  e_0 \in \mathcal{E},\ t=0,1,\ldots,(T-1),
	\end{array}
\end{equation}
where $e_t$, $u_t$ and $w_t(e_t, u_t)$ denote the realized system state, control input and model uncertainty at time step $t$ respectively,  and $(\bar{e}_t, u_t(\bar{e}_t))$ denote the nominal state and corresponding nominal input. Notice that \eqref{eq:generalized_InfOCP} minimizes the nominal cost over a task duration of length $T$ decided by the user, having considered the safety restrictions during an experiment. The cost comprises of the positive definite stage cost $\ell(\cdot, \cdot)$, and the terminal cost $Q(\cdot)$. We point out that, as  system \eqref{eq:unc_system} is uncertain, the optimal control problem \eqref{eq:generalized_InfOCP} consists of finding $[u_0,u_1(\cdot),u_2(\cdot),\ldots]$, where $u_t: \mathbb{R}^{2}\ni x_t \mapsto u_t = u_t(e_t)\in\mathbb{R}^{2}$ are state feedback policies. 

%%%%%%%%%%%%%%%%%%%%%%%%%%%%%%%%%%%%%%%%%%%%
The main challenge in solving problem \eqref{eq:generalized_InfOCP} is that it is difficult to obtain the camera measurement noise distribution support $\mathbb{V}$. 
% This support is needed to calculate the imposed constraint set $\mathcal{E}$ in \eqref{act_esxf}. 
Resorting to worst-case a-priori set estimates of $\mathbb{V}$ as in \cite{tanaskovic2014adaptive, luAccCannon} might result in loss of feasibility of \eqref{eq:generalized_InfOCP}. To avoid this, we use a data-driven estimate of $\mathbb{V}$ denoted by $\hat{\mathbb{V}}(n)$, where $n$ is the number of samples of noise $v_t$ used to construct the set. 
% The corresponding estimate of $\mathcal{E}$ is denoted by $\mathcal{E}$. 
%%%%%%%

\subsection{Control Formulation}\label{sec:iter_cftoc}
As we have noisy output feedback in \eqref{eq:MPC_R_fin}, we follow \cite{mayne2006robust} for a tractable constrained finite time optimal controller design strategy. We repeatedly solve \eqref{eq:generalized_InfOCP} at times $0 \leq t \leq (T-1)$ in a shrinking horizon fashion \cite[Chapter~9]{borrelli2017predictive}. We make the following assumption for this purpose:
%%%%%%%%%%%%%%%%%%%%%%%%%%%%%%%%%%%%%%%%%%%%%
\begin{assumption}\label{assump:origin_con}
The sets $\mathbb{W}_m, \mathbb{W}, \mathbb{V}$, and $\mathcal{U}$ contain the origin in their interior.  
\end{assumption}

%%%%%%%%%%%%%%%%%%%%%%%%%%%%%%%%%%%%%%%%%%%
\subsubsection{Observer Design and Control Policy Parametrization}\label{ssec:obs_cont} 
We design a Luenberger observer for the state as
\begin{align*}
    \hat{e}_{t+1} = A \hat{e}_t + B  u_t + L(y_t-\hat{e}_t),
\end{align*}
where the observer gain $L$ is chosen such that $(A-L)$ is Schur stable. The control policy parametrization for solving \eqref{eq:generalized_InfOCP} is chosen as:
\begin{align*}
     u_t =  \bar{u}_t + K(\hat{e}_t - \bar{e}_t),
\end{align*}
where state feedback policy gain matrix $K$ is chosen such that $(A+BK)$ is Schur stable. 
%%
% \subsubsection{Estimating the Set $\mathcal{E}$}\label{essitimateec}
% % Recall set $\mathcal{E}_S(x_\mathrm{f})$ from Section~\ref{ssec:swingOpt} and \eqref{eq:constraints_nominal}. 
% As $\mathbb{V}$ is unknown in \eqref{bxyxrelation}, $\mathcal{E}$ cannot be computed exactly in \eqref{act_esxf}. Therefore constraints \eqref{eq:constraints_nominal} cannot be imposed while solving \eqref{eq:generalized_InfOCP}. We compute an estimate of $\mathcal{E}$ using the measurement noise support estimate $\hat{\mathbb{V}}(n)$ in \eqref{bxyxrelation}-\eqref{act_esxf} as follows:
% \begin{align}\label{esxf}
%     & {\gamma}^{(i)} = \Vert \mathrm{vert}^{(i)}(y_0 \oplus -(\hat{\mathbb{V}}(n))) \Vert_\infty ,~i \in \{1,2\}, \nonumber \\
%     &\mathcal{E} = \{x: -{\gamma} \leq x \leq {\gamma}\},~\hat{\gamma} = [{\gamma}^{(1)}, {\gamma}^{(2)}]^\top.
% \end{align}
% Note that if $\hat{\mathbb{V}}(n) = \mathbb{V}$, then $\mathcal{E} = \mathcal{E}$. Estimate \eqref{esxf} is used in our control design. 
%%
\subsubsection{Optimal Control Problem}
%%%%%%%%%%%%%%%%%%%%%%%%%%%%%%%%%%
Consider the tightened constraint sets,  
\begin{subequations}\label{eq:tightened_sets}
\begin{align}
    \bar{\mathcal{E}}(n) & = \mathcal{E} \ominus (\mathcal{R}^\mathrm{est}(n) \oplus \mathcal{R}^\mathrm{con}(n)),\\
    \bar{\mathcal{U}}(n) & = \mathcal{U} \ominus K \mathcal{R}^\mathrm{con}(n),
\end{align}
\end{subequations}
%%%%%%%%%%%%%%%%%%%%%%%%%%%%%%
where following \cite[Proposition~1-2]{mayne2006robust}, the set $\mathcal{R}^\mathrm{est}(n)$ is our best estimate of the minimal Robust Positive Invariant set $\mathcal{R}^\mathrm{est}$ for the \emph{estimation} error $\delta e^{\mathrm{est}}_t = e_t - \hat{e}_t$ dynamics defined as
\begin{align}\label{est_R}
    \delta e^{\mathrm{est}}_{t+1} = (A-L) \delta e^{\mathrm{est}}_{t} +  w_t(e_t, u_t) -Lv_t,
\end{align}
and the set $\mathcal{R}^\mathrm{con}(n)$ is our best estimate of the minimal Robust Positive Invariant set $\mathcal{R}^\mathrm{con}$ for the \emph{control} error $\delta e^{\mathrm{con}}_t = \hat{e}_t - \bar{e}_t$ dynamics defined as
\begin{align}\label{con_R}
    \delta e^{\mathrm{con}}_{t+1} = (A+BK) \delta e^{\mathrm{con}}_{t} +  L\delta e^{\mathrm{est}}_{t} + Lv_t,
\end{align}
with $v_t \in \hat{\mathbb{V}}(n)$ and $w_t(e_t, u_t) \in \mathbb{W}$. We use the phrase \emph{best estimate} for the above sets, since  $\hat{\mathbb{V}}(n)$ is an estimate of true and unknown set $\mathbb{V}$. 
% Numerical schemes to construct sets $\mathcal{R}^\mathrm{est}$ and $\mathcal{R}^\mathrm{con}$ under Assumption~\ref{assump:origin_con} can be found in \cite[Section~3.2.2]{kouvaritakis2016model}. 

Using these sets we then solve the following tractable finite horizon constrained optimal control problem at any time step $t \geq 0$ as an approximation to \eqref{eq:generalized_InfOCP}:
\begin{equation} \label{eq:MPC_R_fin}
	\begin{aligned}
	  & V_{t \rightarrow T}^{\star}(\bar{\mathcal{E}}(n), \bar{\mathcal{U}}(n), \mathcal{R}^\mathrm{con}(n), \hat{e}_t)  :=	\\
	& \min_{\bar{e}_{t}, \bar{u}_{t}, \dots,  \bar{u}_{T-1}} ~~ \sum_{k=t}^{T-1} \ell(\bar{e}_{k},  \bar{u}_{k}) + Q(\bar{e}_{T})\\
% 		&~~~~~~\text{s.t., }~~~~~~~ e_{k+1} = Ae_{k} + B  u_{k} + w_{k}(e_{k},  u_{k}),\\
		& ~~~~~~\text{s.t., }~~~~~~~\bar{e}_{k+1} = A\bar{e}_{k} + B\bar{u}_{k},\\
		& ~~~~~~~~~~~~~~~~~~ u_k =  \bar{u}_k + K(\hat{e}_k - \bar{e}_k),\\
		&~~~~~~~~~~~~~~~~~~ \bar{e}_{k} \in \bar{\mathcal{E}}(n),\bar{u}_{k} \in \bar{\mathcal{U}}(n),\\
		&~~~~~~~~~~~~~~~~~~\hat{e}_{t} - \bar{e}_{t} \in \mathcal{R}^\mathrm{con}(n),\\
	    &~~~~~~~~~~~~~~~~~~ \bar{e}_{T} =0,\\
	    & ~~~~~~~~~~~~~~~~~~\forall k \in \{t,t+1,\dots,(T-1)\},
	\end{aligned}
\end{equation}
where $\hat{e}_t$ is the observed state at time step $t$, and $\{\bar{e}_{k}, \bar{u}_{k}\}$  denote the nominal state and corresponding input respectively predicted at time step $k \geq t$. 
% The controller minimizes the cost over the predicted nominal trajectory $\Big \{ \{\bar{e}_{k}, \bar{u}_{k}\}_{k=t}^{T-1}, \bar{e}_{T} \Big \}$, which comprises of the positive definite stage cost $\ell(\cdot, \cdot)$, and the terminal cost $Q(\cdot)$ as defined in \eqref{eq:generalized_InfOCP}.  
After solving \eqref{eq:MPC_R_fin}, in closed-loop we apply
\begin{equation}\label{eq:inputCL_DF}
	u^{\star}_t(e_t) : u^{\star}_t = \bar{u}_{t}^{\star} + K(\hat{e}_t - \bar{e}^{\star}_t)
\end{equation}
to system \eqref{eq:unc_system}. We then resolve the problem \eqref{eq:MPC_R_fin} again at the next $(t+1)$-th time step, yielding a shrinking horizon strategy. The choice of initial observer state is made as follows: 
\begin{align}\label{polXH}
\hat{e}_0  \in  - (\mathcal{R}^\mathrm{est}(n) \ominus \mathcal{E}).
\end{align}

\begin{assumption}[Manipulator Speed]\label{assump:reach_ball}
If any feasible solution is found to \eqref{eq:MPC_R_fin} satisfying velocity error constraints $\bar{\mathcal{U}}(n)$, the manipulator has enough velocity authority to satisfy these constraints, where the predicted ball velocity is obtained using forward Euler integration at free-fall. 
\end{assumption}

Recall the set $\mathcal{E}_\mathrm{tr}$ containing the set of all possible errors $e_0$ at the start of the catch phase, shown in Fig.~\ref{fig:roll_out}. We now make the following assumption. 
\begin{assumption}[Well Posedness]\label{ass:well}
We assume that given state $e_0 \in \mathcal{E}_\mathrm{tr}$, optimization problem \eqref{eq:MPC_R_fin} is feasible at all time steps $0 \leq t \leq (T-1)$ with model uncertainty support ${\mathbb{W}}$, and true measurement noise support $\hat{\mathbb{V}}(n) = \mathbb{V}$ used in \eqref{est_R}-\eqref{con_R} and \eqref{polXH}, when \eqref{eq:inputCL_DF} is applied to \eqref{eq:unc_system} in closed-loop. This implies that $e_t \in \mathcal{E}$ for all $0 \leq t \leq T$, where $\mathcal{E}$ is obtained from $\mathcal{E}_\mathrm{tr}$ following \eqref{act_esxf}.
\end{assumption}
\begin{definition}[Trial Failure]\label{def:stateConstraintFailure} A Trial Failure at time step $t$ is the event 
\begin{align*}
   \mathrm{[TF]}_t:~ e_t \notin \mathcal{E},~ 0 \leq t \leq T.
\end{align*}
That is, a Trial Failure implies the violation of imposed constraints \eqref{eq:constraints_nominal} by system \eqref{eq:unc_system} in closed-loop with feedback controller \eqref{eq:inputCL_DF}. 
% This is highlighted in Fig.~\ref{fig:IF}.
\end{definition}
% \begin{figure}[h]
%     \includegraphics[width=\columnwidth]{canDama.jpg}
%     \caption{Illustration of Iteration Failure Limits During Catch Phase}
%     \label{fig:IF}
% \end{figure}
Note that a Trial Failure is a possible scenario only because $\mathbb{V}$ is unknown and is estimated with $\hat{\mathbb{V}}(n)$ in \eqref{eq:MPC_R_fin}. 
% Let $T_\mathrm{tf} < T$ be the time step when a Trial Failure occurs. In that case, problem \eqref{eq:MPC_R_fin} becomes infeasible at $T_\mathrm{tf}$. We then stop the trial and proceed to the next attempt. 
Intuitively, a Trial Failure implies one of the following:
\begin{enumerate}[(P1)]
  \item Problem \eqref{eq:MPC_R_fin} losing feasibility during $0 < t < T$. This happens if $\hat{\mathbb{V}}(n) \not \supset \mathbb{V}$. 
 
%  \item The string becoming taut during the catch-phase, where modeling \eqref{eq:unc_system} becomes invalid. This can happen if $\hat{\mathbb{V}}(n) \supset \mathbb{V}$, and thus $\mathcal{E} \supset \mathcal{E}$.  
 
 \item Problem \eqref{eq:MPC_R_fin} losing feasibility initially at $t=0$, and/or sets $\bar{\mathcal{E}}(n), \bar{\mathcal{U}}(n)$ becoming empty. This can happen if $\hat{\mathbb{V}}(n) \supset \mathbb{V}$.
\end{enumerate}
 
% The probability of Trial Failure $[\mathrm{TF}]_t$ is a function of the set $\hat{\mathbb{V}}(n)$. 
% % However, it is not enough to focus on probability of $[\mathrm{TF}]_t$ alone. 
% For example, a low probability of $[\mathrm{TF}]_t$ 
% % before the ball coming to contact with the cup 
% can be achieved by considering worst-case apriori estimates for $\mathbb{V}$, but then it may result in infeasibility of \eqref{eq:MPC_R_fin} at time step $t=0$, or sets $\bar{\mathcal{E}}(n)$ or $\bar{\mathcal{U}}(n)$ being empty. 
% In the next section, we introduce two design specifications (D1) and (D2) to formalize this joint focus. We then show how the sets $\hat{\mathbb{V}}(n)$ are systematically constructed and updated.

%%%%%%%%%%%%%%%%%%%%%%%%%%%%%%%%%%%%%%%%%%%%%%%
\subsection{Constructing Set $\hat{\mathbb{V}}(n)$}\label{sec:lrbf}
As described in Section~\ref{ssec:feedback_for} the set $\hat{\mathbb{V}}(n)$ is an estimate of the measurement noise support $\mathbb{V}$, derived from $n$ samples of noise $v_t$. The set $\hat{\mathbb{V}}(n)$ is then used to compute $\mathcal{R}^\mathrm{est}(n)$ and $\mathcal{R}^\mathrm{con}(n)$ in \eqref{est_R}-\eqref{con_R}, used in \eqref{eq:MPC_R_fin} and \eqref{polXH}. We consider the following two design specifications while constructing set $\hat{\mathbb{V}}(n)$, given a fixed sample size $n$. 
\begin{enumerate}[(D1)]
    \item Probability of the event $\hat{\mathbb{V}}(n) \not \supset \mathbb{V}$ is bounded with a user specified upper bound $\epsilon$. 
    % \item Closed-loop control law \eqref{eq:inputCL_DF} ensures that system \eqref{eq:unc_system} satisfies a user specified upper bound $\epsilon$ on probability of the event (P1) in Trial Failure,
    \item Estimate $\hat{\mathbb{V}}(n)$ ensures event (P2) in Trial Failure occurs with a vanishing probability, while satisfying specification (D1).
\end{enumerate}
\subsubsection*{Satisfying (D1) using Distribution Information}
\begin{figure}[h]
    \includegraphics[width=\columnwidth]{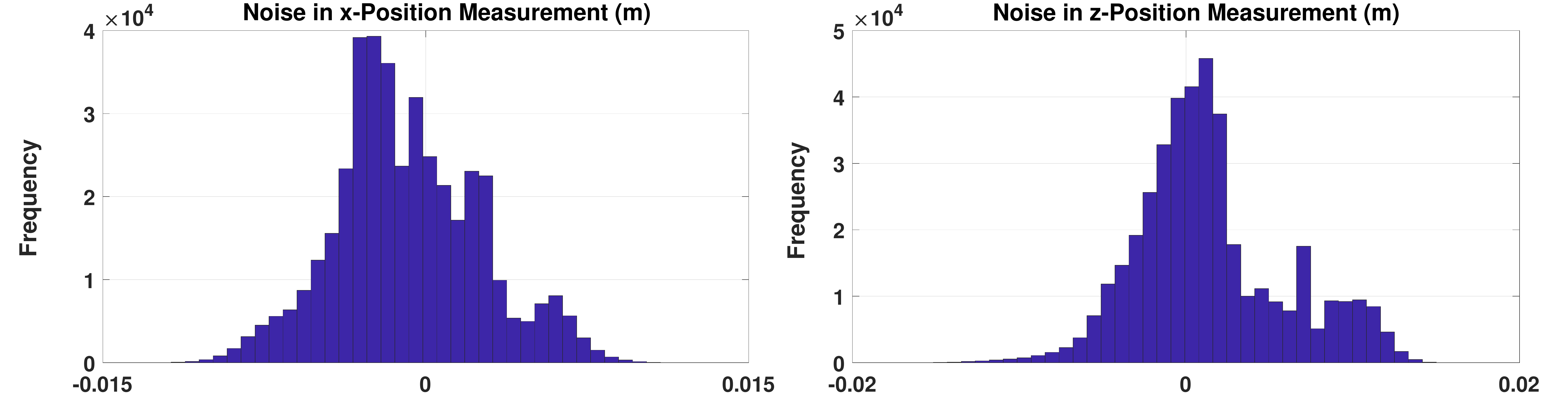}
    \caption{Camera measurement noise distribution histogram for a fixed camera environment using $n = 400,000$ samples.}
    \vspace{-8pt}
    \label{fig:estim_con2}
\end{figure}

Fig.~\ref{fig:kendama_intro} shows the configuration of the system when $n$ noise samples are collected to construct $\hat{\mathbb{V}}(n)$. Let Assumption~\ref{assump:end_eff_acc} hold true and the ball is held still, vertically below the end-effector at a position, whose $z$-coordinate $z^\mathrm{cup} = \bar{z}$ is fixed and \emph{known} from previous UR5e end-effector measurements, and $x$-coordinate is fixed at $x^\mathrm{ball} = 0$. 
% The time spent in this configuration before the swing-up phase of an iteration is decided by the user, and say is denoted by $t_\mathrm{con}$. 
We then collect $n$ camera position measurements of the ball at this configuration. The discrepancy between the known position and the measurements yield values of noise samples $\mathbf{v}_n = [v_0, v_1, \dots, v_n]$. For a fixed \emph{environment},\footnote{camera environment is parametrized by say lighting conditions, camera field of view, etc.} the distribution of collected samples is shown in Fig.~\ref{fig:estim_con2}, which is approximately a truncated normal distribution.
We thereby consider this distribution family in Fig.~\ref{fig:estim_con2} conditioned on any environment as 
\begin{align}\label{eq:dist_fam}
    \mathcal{P}^q_{\theta_q} | \mathrm{env} = \mathcal{N}_\mathrm{trunc}(\mu_q,\sigma^2_q,3),~\textrm{with } q \in \{1,2\},
\end{align}
where $\mathcal{P}_\theta$ denotes that the distribution $\mathcal{P}$ belongs to a parametric family (truncated normal) parametrized by $\theta = (\mu, \sigma)$, $q$ denotes the $q^\mathrm{th}$ dimension ($x$ and $z$ directions), and parameters $(\mu_q, \sigma_q)$ are unknown. For a parametric distribution such as \eqref{eq:dist_fam},
% We construct Bootstrap confidence intervals of for the parameters $(\mu_q, \sigma_q)$ for $q \in \{1,2\}$ by re-sampling the collected $\mathbf{v}_n$ samples $n_B$ number of times. 
for any chosen $\epsilon \in (0,1)$,  set $\hat{\mathbb{V}}(n)$ is then constructed  as the $(1-\epsilon)$-Confidence Support of $\mathcal{P}_\theta | \mathrm{env}$ using the method in \cite{bujarbaruah2019learning}, which ensures 
\begin{align}\label{eq:noise_fail}
    \mathbb{P}(\hat{\mathbb{V}}(n) \not \supset \mathbb{V} ) \leq \epsilon. 
    % \implies \mathbb{P}(\mathcal{E} \not \subseteq \mathcal{E}) \leq \epsilon.
\end{align}
Note that \eqref{eq:noise_fail} is a sufficient condition to guarantee that if (D2) holds, solving \eqref{eq:MPC_R_fin} and applying \eqref{eq:inputCL_DF} to \eqref{eq:unc_system} gives
\begin{align}\label{eq:suff_condition_failure}
    % \mathbb{P}(v_t \notin \hat{\mathbb{V}}(n)) \leq \epsilon \implies 
    \mathbb{P}(e_t \notin {\mathcal{E}}) \leq \epsilon,~0\leq t \leq T,
\end{align}
if $\hat{\mathbb{V}}(n)$ is used to construct sets $\mathcal{R}^\mathrm{est}(n)$ and $\mathcal{R}^\mathrm{con}(n)$.
%%%%%%%%%%%%%%%%%%%%%%%%%%%%%%%%%%%
\subsubsection*{Satisfying (D2) using Assumption~\ref{ass:well}}

Since Assumption~\ref{ass:well} holds, there exists a number of noise samples $n_\epsilon$ for any $\epsilon \in (0,1)$, such that $\hat{\mathbb{V}}(n_\epsilon)$ satisfies (D2). Thus, only the sample size $n$ has to be chosen\footnote{for $n$ fixed, $\epsilon$ can be increased while constructing $\hat{\mathbb{V}}(n)$ to satisfy (D2).} for $\hat{\mathbb{V}}(n)$ appropriately to satisfy (D2), having ensured \eqref{eq:suff_condition_failure}. This guarantees that constructing sets $\mathcal{R}^\mathrm{con}(n)$ and $\mathcal{R}^\mathrm{est}(n)$ using $\hat{\mathbb{V}}(n)$ and then designing a feedback control by  solving \eqref{eq:MPC_R_fin} results in problem \eqref{eq:MPC_R_fin} being feasible throughout the task with probability at least $\beta = (1-\epsilon)^{T-1}$. Value of $\epsilon$ can be chosen small enough for any user-specified level $\beta$ can be attained.
%%%%
\subsection{Obtaining Catches}\label{wellsec}
Constructing $\hat{\mathbb{V}}(n)$ as per Section~\ref{sec:lrbf} to ensure \eqref{eq:suff_condition_failure} is still \emph{not} a sufficient condition to obtain a catch in an experiment with specified probability $\beta$, as our model \eqref{eq:unc_system} does not account for additional factors such as object dimensions, presence of contact forces, etc. 

To that regard, we introduce the notion of a \emph{successful catch}, which is defined as the ball successfully ending up inside the cup at the end of a roll-out. Thus, a successful catch accounts for the dimensions of the ball and the cup, and the presence of contact forces. 

\begin{assumption}[Existence of a Successful Catch]\label{assump:swingup_end}
We assume that given an initial state $e_0 \in \mathcal{E}_\mathrm{tr}$, an input policy obtained by solving \eqref{eq:MPC_R_fin} can yield a  successful catch, if true measurement noise support $\mathbb{V}$ were known \emph{exactly}. 
\end{assumption}
\begin{remark}\label{rem_asym}
% From \cite{bujarbaruah2019learning} we know that as long as confidence intervals for parameters $(\mu, \sigma)$ in \eqref{eq:dist_fam} converge, $\hat{\mathbb{V}}(n) \rightarrow \mathbb{V}$ as $n \rightarrow \infty$. 
From \cite{bujarbaruah2019learning} we know that as long as confidence intervals for parameters $(\mu, \sigma)$ in \eqref{eq:dist_fam} converge, $\hat{\mathbb{V}}(n) \rightarrow \mathbb{V}$ as $n \rightarrow \infty$. So, if sample size $n$ is increased iteratively approaching $n \rightarrow \infty$, obtaining a successful catch guaranteed  owing to Assumption~\ref{assump:swingup_end}. However if a precise positioning system like Vicon is used to collect the noise samples, due to limited access to such environments, collecting more samples and increasing $n$ could be expensive. We therefore stick to our method of constructing $\hat{\mathbb{V}}(n)$ for a \emph{fixed} $n$ as per Section~\ref{sec:lrbf}, and we attempt successful catches with multiple roll-outs by solving \eqref{eq:MPC_R_fin}. For improving the empirical probability of successful catches in these roll-outs, one may then increase $n$ and thus update the control policy. We demonstrate this in Section~\ref{ssec:learn_catch}. 
\end{remark}

%%%%%%%%%%%%%%%%%%%%%%%%%%%%%%
\section{Experimental Results}\label{sec:exp}
We present our preliminary experimental findings in this section. For our experiments, the original Kendama handle was modified to be attached to a 3D printed mount on the UR5e end-effector, as shown in Fig.~\ref{fig:kendama_intro}. A single Intel RealSense D435 depth camera running at 60 FPS was used to estimate the position and velocity of the ball.

\subsection{Control Design in the Catch Phase}
Once the swing-up controller is designed as per Section~\ref{ssec:swingOpt} and an open-loop swing-up control sequence is applied to the manipulator, we design the feedback controller by finding approximate solutions to the following problem:
\begin{equation}\label{example}
	\begin{array}{clll}
% 		& V^{j,\star}(e^j_0)  = \\ [1ex]
% % 		\hspace{-0.cm}    & V^{j,\star}(x_S)  =
		& \displaystyle\min_{u_0,u_1(\cdot),\ldots}  \displaystyle\sum\limits_{t=0}^{T-1} 500\left \| \bar{e}_t \right\|^2_2 + 0.4 \left\| u_t(\bar{e}_t) \right\|^2_2  \\[1ex]
		& ~~~~\text{s.t.,}\\
		& ~~~~~~~~~~~~ e_{t+1} = Ae_t + Bu_t(e_t),\\
		& ~~~~~~~~~~~~ \bar{e}_{t+1} = A \bar{e}_t + B u_t(\bar{e}_t),\\
		& ~~~~~~~~~~~~y_{t} = e_t + v_t,\\
		& ~~~~~~~~~~~~ {e}_{t} \in {\mathcal{E}} ,~\begin{bmatrix}  -8\textrm{m/s} \\-8\textrm{m/s}
		\end{bmatrix} \leq u_t(e_t)
		 \leq \begin{bmatrix} 8\textrm{m/s} \\8\textrm{m/s}
		\end{bmatrix},\\
		& ~~~~~~~~~~~~ \forall v_t \in \mathbb{V}, \\[0.5ex]
		& ~~~~~~~~~~~~ t=0,1,\ldots,(T-1),
	\end{array}
\end{equation}
% For our simulations, we let the system complete an iteration despite Iteration Failure in order to speedup convergence of $\hat{\mathbb{W}}^j$ to $\mathbb{W}$ (refer to Remark~\ref{rem:softCon}). 
where set $\mathcal{E}_\mathrm{tr} = [-0.316\textrm{m}, 0.349\textrm{m}] \times [-0.2095\textrm{m}, 0.2457\textrm{m}]$, shown in Fig.~\ref{fig:roll_out}. Note that for this specific scenario the presence of model uncertainty can be ignored. Set $\mathbb{V}$ is unknown, and we consider Assumption~\ref{ass:well} holds. System matrices $A, B$ are from Section~\ref{ssec:feedback_for}. We find solutions to \eqref{example} for $T=50~\text{steps}$, i.e., $0.5$ seconds. 
% The feedback gain $K$ in \eqref{eq:inputCL_DF} is chosen to be the optimal LQR gain for system $e^+ = (A+BK)e$ with parameters $Q_\mathrm{LQR}=50{I}_2$ and $R_\mathrm{LQR} = 0.1$. 

% \subsection{Construction of $\hat{\mathbb{V}}(n)$}\label{ssec:constr_v_exp}

%%%%%%%%%%%%%%%%%%%%%%%%%%%%%%%%%%%%%%%%%%%%%%%

\subsection{Learning to Catch}\label{ssec:learn_catch}
\begin{figure}[h]
    \centering
    \includegraphics[width=0.9\columnwidth]{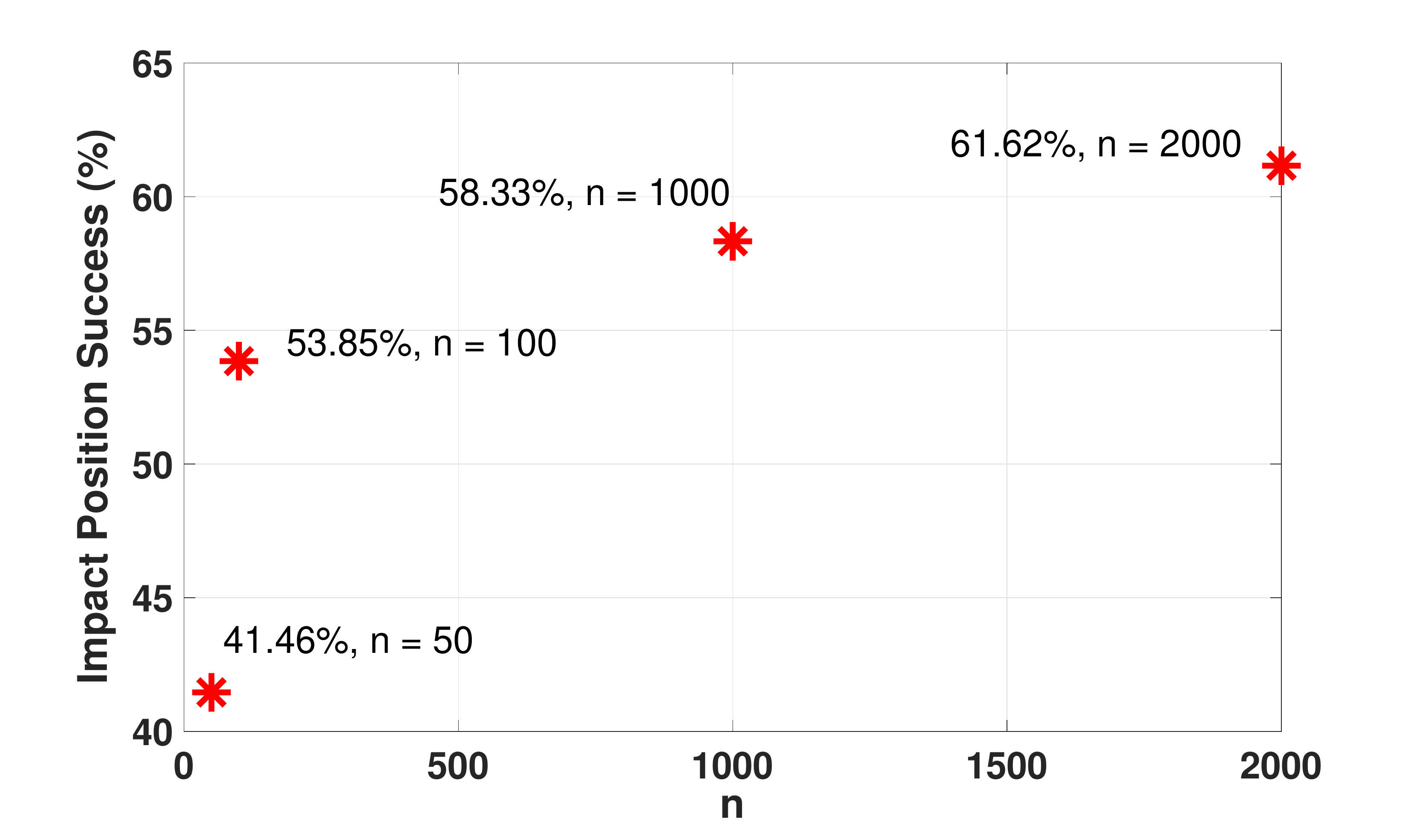}
    \caption{Percentage of times the ball hitting the cup center among all roll-outs vs sample size $n$.}
    \vspace{-8pt}
    \label{fig:hit_center}
\end{figure}

\begin{figure}[h]
    \centering
    \includegraphics[width=0.9\columnwidth]{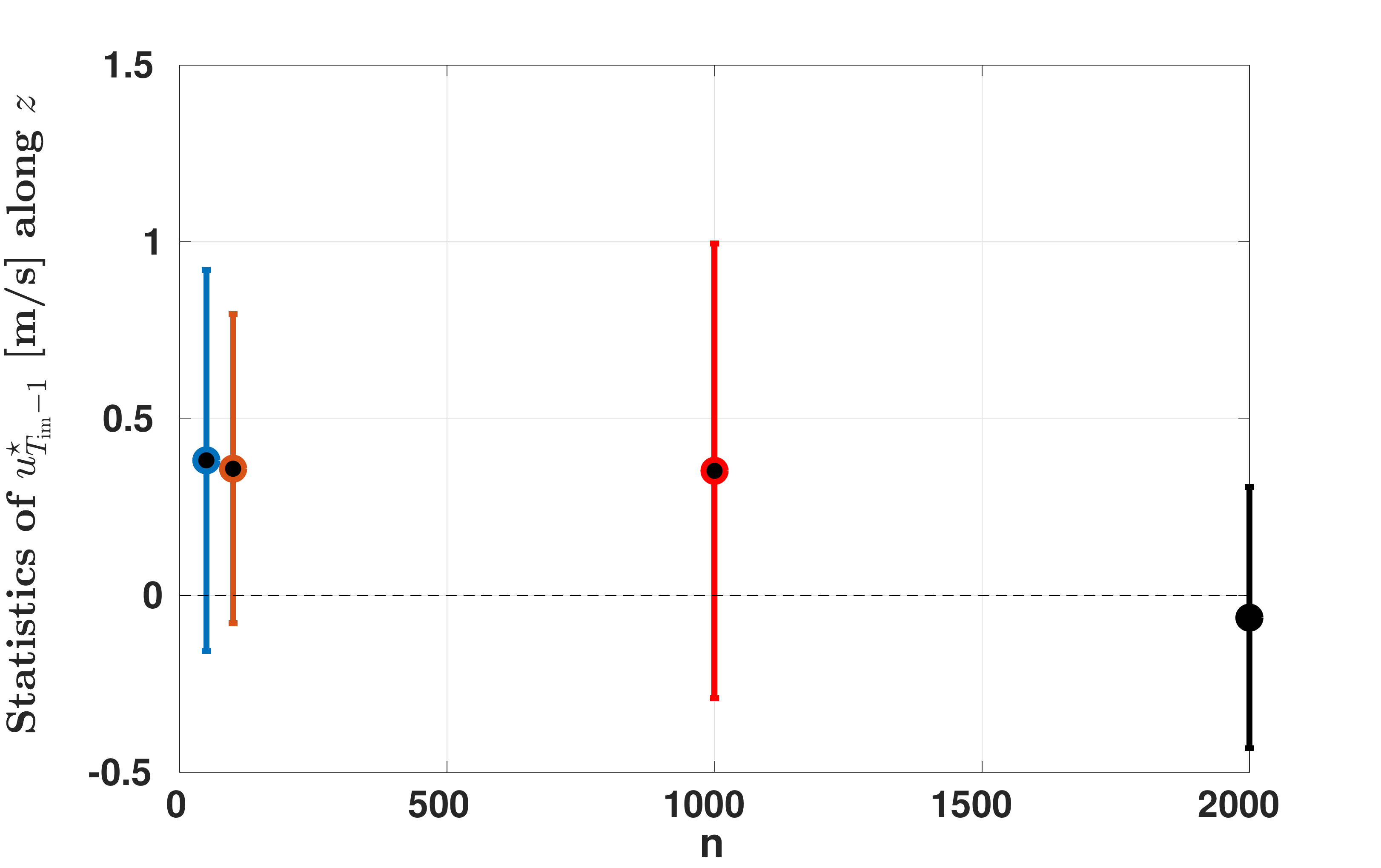}
    \caption{One standard deviation interval around the mean (circle) of $z$-relative velocity at impact, i.e., $[u^\star_{T_\mathrm{im}-1}]_z$ vs sample size $n$.}
    \vspace{-8pt}
    \label{fig:rel_vel}
\end{figure}

% In order to prove iterative catch improvement with updates of the set $\hat{\mathbb{V}}^j$, 
We conduct $50$ \emph{roll-outs} of the catching task by solving \eqref{eq:MPC_R_fin}, having formed $\hat{\mathbb{V}}(n)$ as per Section~\ref{sec:lrbf}, with $n=100$ and then iteratively increasing to $n=2000$. Sets $\hat{\mathbb{V}}(n)$ are formed using \cite{bujarbaruah2019learning}. Fig.~\ref{fig:hit_center} shows the percentage of roll-outs conducted for each iteration (i.e., for each value of $n$), that resulted in the ball successfully striking the center of the cup.  
The percentage increases from $41.46\%$ to $61.62\%$. 
Furthermore, another crucial quantity at the time of impact is the commanded relative velocity \eqref{eq:inputCL_DF} in $z$-direction, a lower value of which indicates an increased likelihood of the ball not bouncing out. The average value and the standard deviation of of $(u^{\star}_{T_\mathrm{im}-1})_z^{*\tilde{m}}$ for $\tilde{m} \in \{1,2,\dots,50\}$ is shown in Fig.~\ref{fig:rel_vel}, where $(\cdot)^{*\tilde{m}}$ denotes the $\tilde{m}^\mathrm{th}$ roll-out and $T_\mathrm{im} \leq T$ denotes the time of impact.  
As seen in Fig.~\ref{fig:rel_vel}, the mean of the relative velocity at impact lowers from $0.38$ m/s to $-0.06$ m/s. This together with Fig.~\ref{fig:hit_center} indicates a possibility of increasing successful catch counts as $n$ is increased.

\subsection{Increasing Successful Catches}
%%%%%%%%%%%%%%%%%%%%%%%%%%%%%%%%%%
In order to prove that the trend shown in Fig.~\ref{fig:hit_center} and Fig.~\ref{fig:rel_vel} results in an increasing number of successful catches, we resort to exhaustive Mujoco \cite{todorov2012mujoco, tassa2020dm_control} simulations\footnote{due to unavailability of laboratory access}. The task duration in this case is $T=25~\text{steps}$. 
\begin{figure}[h]
    \centering
    \includegraphics[width=0.9\columnwidth]{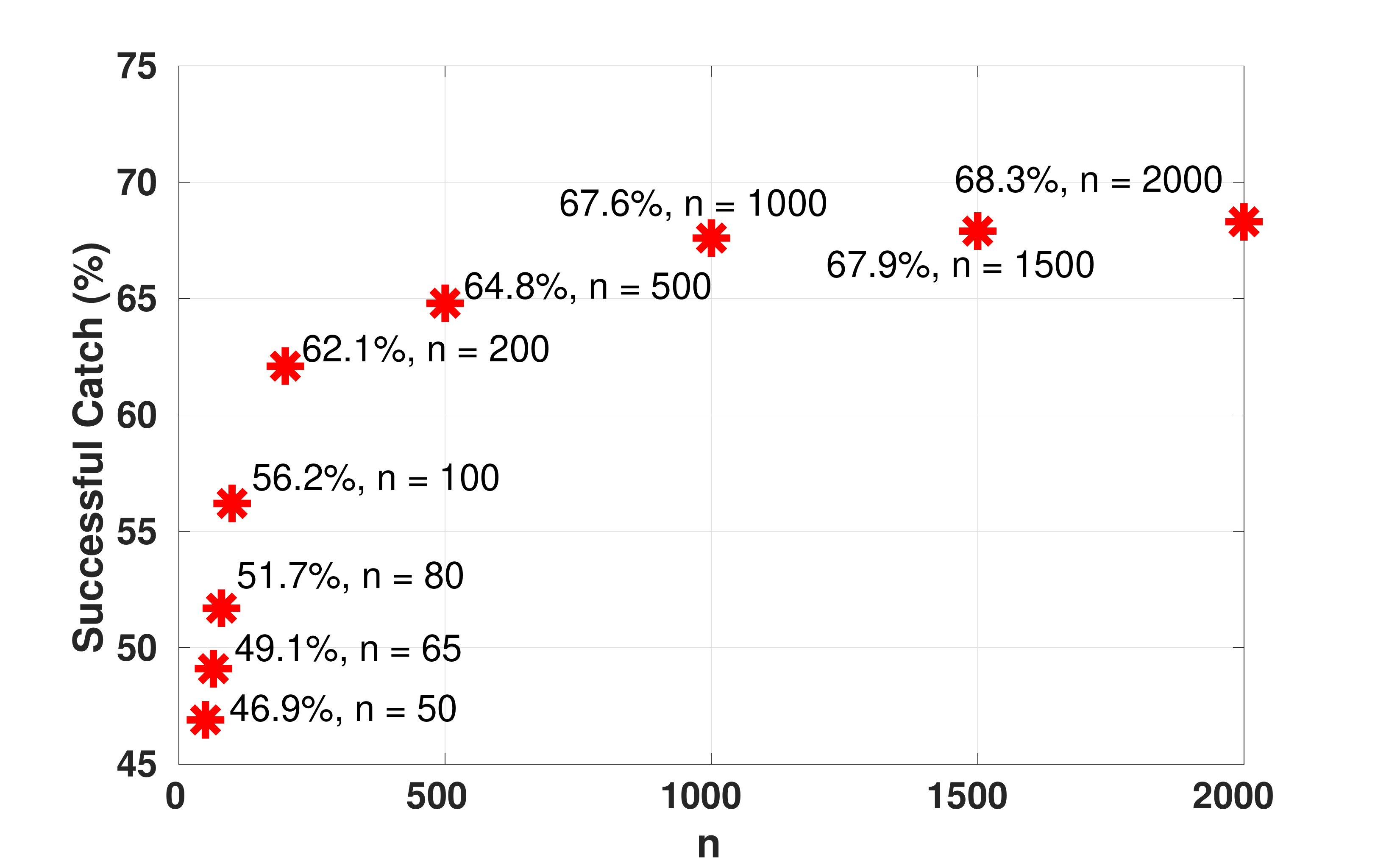}
    \caption{Percentage of successful catches vs sample size $n$.}
    \vspace{-8pt}
    \label{fig:catch_suc_ugoco}
\end{figure}
The trend in the percentage of successful catches with 1000 roll-outs corresponding to each $n$, varying from $n=50$ to $n=2000$, is shown in Fig.~\ref{fig:catch_suc_ugoco}. For $n=50$, $46.9\%$ of the roll-outs result in a successful catch. The number increases to $68.3 \%$ for $n=2000$. This verifies that the preliminary experimental results from Fig.~\ref{fig:hit_center} and Fig.~\ref{fig:rel_vel} would very likely result in a similar trend as in Fig.~\ref{fig:catch_suc_ugoco}. Thus we prove that our proposed approach enables successful learning of the kendama ball catching task.  
%%%%%%%%%%%%%%%%%%%%%%%%%%%%%%%%%%%
\section{Conclusions}\label{sec: concl}
We proposed a model based control strategy for the classic cup-and-ball game. 
% The control problem was divided into two sub-tasks, for swinging the ball up, and then catching the free-falling ball respectively. The swing-up trajectory ensures that a successful catch is possible with our feedback control design approach. Subsequently, a convex optimization problem was solved online during the ball's free-fall to control the manipulator and catch the ball. 
The controller utilized noisy position measurements of the ball from a camera, and the support of this noise distribution was iteratively learned from data. Thus, the closed-loop control policy iteratively updates. We proved that the probability of a catch increases in the limit, as the learned support  nears the true support of the camera noise distribution. Preliminary experimental results and high-fidelity simulations support our analysis.

% Note that the proposed methodology in the paper is general and not restricted to only the cup-and-ball problem. Provided the availability of samples and the validity of the assumptions, our approach can be used to learn and update the distribution estimate of any disturbance in a system for iterative refinement of control policy.  
%%%%%%%%%%%%%%%%%%%%%%%%%%%%%%%%%%%
\section*{Acknowledgement}
We thank Yuri Glauthier, Charlott Vallon, and Sangli Teng for their contributions on the hardware experiments, as well as Vijay Govindarajan, Siddharth Nair and Edward Zhu for extremely useful reviews and discussions. 
The research was funded by grants ONR-N00014-18-1-2833, NSF-1931853, and Siemens.

% \section*{Appendix}
% \subsection{System Matrices and Dynamics}
% The matrices in \eqref{lagrange} are given by,
% \begin{align*}
% &   \mathbf{q}_j=
%     \begin{bmatrix}
%         x^\mathrm{cup}_j \\
%         z^\mathrm{cup}_j \\
%         \phi_j
%     \end{bmatrix}, ~ F_j=[F^x_j , F^z_j, 0]^\top,~ G(\mathbf{q}_j) = 
%     \begin{bmatrix}
%         0 \\
%         (m_c+m_b)g \\
%         m_bgr\sin\phi_j
%     \end{bmatrix},\\
% & C(\mathbf{q}_j,\dot{\mathbf{q}}_j)=
%     \begin{bmatrix}
%         0 & 0 & -m_br\dot{\phi}_j\sin\phi_j \\
%         0 & 0 & m_br\dot{\phi}_j\cos\phi_j \\
%         0 & 0 & 0
%     \end{bmatrix}, \\
% &  M(\mathbf{q}_j)=
%     \begin{bmatrix}
%         (m_c+m_b) & 0 & m_br\cos\phi_j \\
%         0 & (m_c+m_b) & m_br\sin\phi_j \\
%         m_br\cos\phi_j & m_br\sin\phi_j & m_br^2 
%     \end{bmatrix},
% \end{align*}
% for all $j \geq 0$. The state space form of \eqref{lagrange} in \eqref{statespacelagrange} is given by
% \begin{align*} 
%     \begin{bmatrix}
%     \dot{\mathbf{q}}_j\\
%     \ddot{\mathbf{q}}_j
%     \end{bmatrix} 
%     & =
%     \begin{bmatrix}
%     \mathbf{0}_{3 \times 3} && {I}_3 \\
%     \mathbf{0}_{3 \times 3} && -M^{-1}(\mathbf{q}_j)C(\mathbf{q}_j,\dot{\mathbf{q}}_j)
%     \end{bmatrix}
%     \begin{bmatrix}
%     \mathbf{q}_j\\
%     \dot{\mathbf{q}}_j
%     \end{bmatrix} \\ 
%     & ~~~~~~~~~~~~ +
%     \begin{bmatrix}
%     \mathbf{0}_{3 \times 1}\\
%     M^{-1}(\mathbf{q}_j)\big(-G(\mathbf{q}_j)+F_j\big)
%     \end{bmatrix},\\
%     & = f(\begin{bmatrix} \mathbf{q}_j\\ \dot{\mathbf{q}}_j\end{bmatrix}, F_j),~\forall j \geq 0.
%  \end{align*}
%%%%%%%%%%%%%%%
%\balance
\renewcommand{\baselinestretch}{0.8}
%%%%%%%%%%%%%%%%%%%%%%%%%%%%%%%
\bibliographystyle{IEEEtran} 
\bibliography{IEEEabrv,bibliography.bib}
\end{document}